\documentclass[letterpaper, 10 pt, conference]{ieeeconf}  

\IEEEoverridecommandlockouts                              

\usepackage{url}
\usepackage{subfig}
\usepackage{graphicx}
\usepackage{varioref}
\graphicspath{{figures/}}
\labelformat{equation}{\textup{(#1)}}
\usepackage[pageanchor=true,plainpages=false,pdfpagelabels,bookmarks,bookmarksnumbered]{hyperref}
\usepackage{amssymb}
\usepackage{multirow}
\usepackage{todonotes}
\usepackage{flushend}

\usepackage{array}
\newcolumntype{L}[1]{>{\raggedright\let\newline\\\arraybackslash\hspace{0pt}}m{#1}}
\newcolumntype{C}[1]{>{\centering\let\newline\\\arraybackslash\hspace{0pt}}m{#1}}
\newcolumntype{R}[1]{>{\raggedleft\let\newline\\\arraybackslash\hspace{0pt}}m{#1}}

\def\usenatbib{1}
\ifx\usenatbib\undefined
    \usepackage{cite}
\else
    \makeatletter
    \let\NAT@parse\undefined
    \makeatother
    \usepackage[numbers,sort]{natbib}

    \setcitestyle{citesep={], [}}
    \makeatletter
    \def\NAT@def@citea{\def\@citea{\NAT@separator}}%
    \makeatother
\fi

\graphicspath{ {./figures/} }

\let\orgautoref\autoref
\providecommand{\Autoref}
        {\def\equationautorefname{Equation}%
         \def\figureautorefname{Figure}%
         \def\subfigureautorefname{Figure}%
         \def\Itemautorefname{Item}%
         \def\tableautorefname{Table}%
         \def\exerciseautorefname{Exercise}%
         \def\starexerciseautorefname{Exercise}%
         \def\sectionautorefname{Section}%
         \def\subsectionautorefname{Section}%
         \def\subsubsectionautorefname{Section}%
         \def\chapterautorefname{Section}%
         \def\partautorefname{Part}%
         \orgautoref}

\renewcommand{\autoref}
        {\def\equationautorefname{Equation}%
         \def\figureautorefname{Fig.}%
         \def\subfigureautorefname{Fig.}%
         \def\Itemautorefname{item}%
         \def\tableautorefname{Table}%
         \def\exerciseautorefname{Exercise}%
         \def\starexerciseautorefname{Exercise}%
         \def\sectionautorefname{Section}%
         \def\subsectionautorefname{Section}%
         \def\subsubsectionautorefname{Section}%
         \def\chapterautorefname{Section}%
         \def\partautorefname{Part}%
         \orgautoref}

\usepackage{hyperref}
\hypersetup{
colorlinks=true,
urlcolor=red,
}
\title{\LARGE \bf
DAVIS-Ag: A Synthetic Plant Dataset for Prototyping\\ Domain-Inspired Active Vision in Agricultural Robots
}

\author{
Taeyeong Choi$^{1}$, Dario Guevara$^{2}$, Zifei Cheng$^{3}$, Grisha Bandodkar$^{3}$, \\
Chonghan Wang$^{3}$, Brian N. Bailey$^{4}$, Mason Earles$^{2,5}$, and Xin Liu$^{3}$ 
\thanks{
$^{1}$Department of Information Technology at Kennesaw State University, USA, 
$^{2}$Department of Viticulture and Enology, 
$^{3}$Department of Computer Science, 
$^{4}$Department of Plant Sciences, and
$^{5}$Department of Biological and Agricultural Engineering at the University of California, Davis, USA. 
  {\tt\small tchoi3@kennesaw.edu}, 
  {\tt\small \{dguevara, zfcheng, gbandodkar, wchwang, bnbailey, jmearles, xinliu\}@udcavis.edu}
  } 
}

\begin{document}

\onecolumn
\vspace*{\fill}
    © 2024 IEEE. Personal use of this material is permitted. Permission from IEEE must be obtained for all other uses, in any current or future media, including reprinting/republishing this material for advertising or promotional purposes, creating new collective works, for resale or redistribution to servers or lists, or reuse of any copyrighted component of this work in other works.
\vspace*{\fill}
\twocolumn
\clearpage

\maketitle
\thispagestyle{empty}
\pagestyle{empty}

\begin{abstract}

  In agricultural environments, viewpoint planning can be a critical functionality for a robot 
with visual sensors to obtain informative observations of objects of interest (e.g.,~fruits) 
from complex structures of plant with random occlusions. 
Although recent studies on active vision have shown some potential for agricultural tasks, 
each model has been designed and validated on a unique environment that would not easily 
be replicated for benchmarking novel methods being developed later. 
In this paper, we introduce a dataset, so-called \mbox{DAVIS-Ag}, for promoting more extensive research on Domain-inspired Active VISion in Agriculture. 
To be specific, we leveraged our open-source ``AgML''~framework and $3$D~plant simulator of ``Helios'' to produce $502$K~RGB images from $30$K~densely sampled spatial locations in $632$~synthetic orchards.
Moreover, plant environments of strawberries, tomatoes, and grapes are considered at two different scales (i.e.,~Single-Plant and Multi-Plant).
Useful labels are also provided for each image, including~(1)~bounding boxes and 
(2)~instance segmentation masks for all identifiable fruits, and also (3)~pointers to other images of the viewpoints that are \emph{reachable} by an execution of action so as to simulate active viewpoint selections at each time step. 
Using~\mbox{DAVIS-Ag}, we visualize motivating examples where 
fruit visibility can dramatically change depending on the pose of the camera view primarily due to occlusions by other components, such as leaves. 
Furthermore, we present several baseline models with experiment results for benchmarking in the task of target visibility maximization.
Transferability to real strawberry environments is also investigated to demonstrate the feasibility of using
the dataset for prototyping real-world solutions.
For future research, our dataset is made publicly available online: \url{https://github.com/ctyeong/DAVIS-Ag}.

\end{abstract}

\section{INTRODUCTION}

\begin{figure}[t]\centering
    \subfloat[]{\label{fig:single_vine}%
    \includegraphics[width=.47\linewidth]{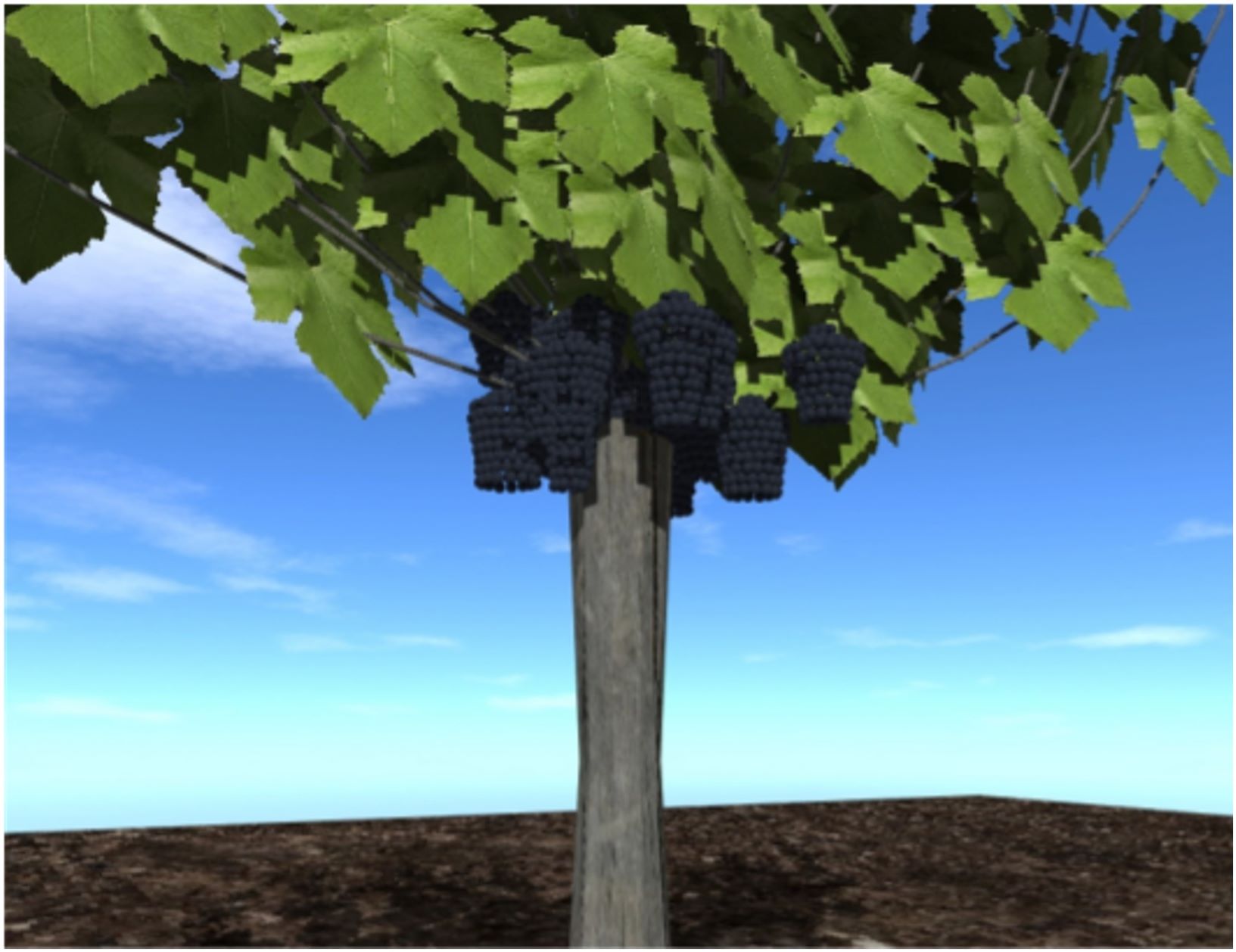}}
    \quad
    \subfloat[]{\label{fig:single_vine_seg}%
    \includegraphics[width=.47\linewidth]{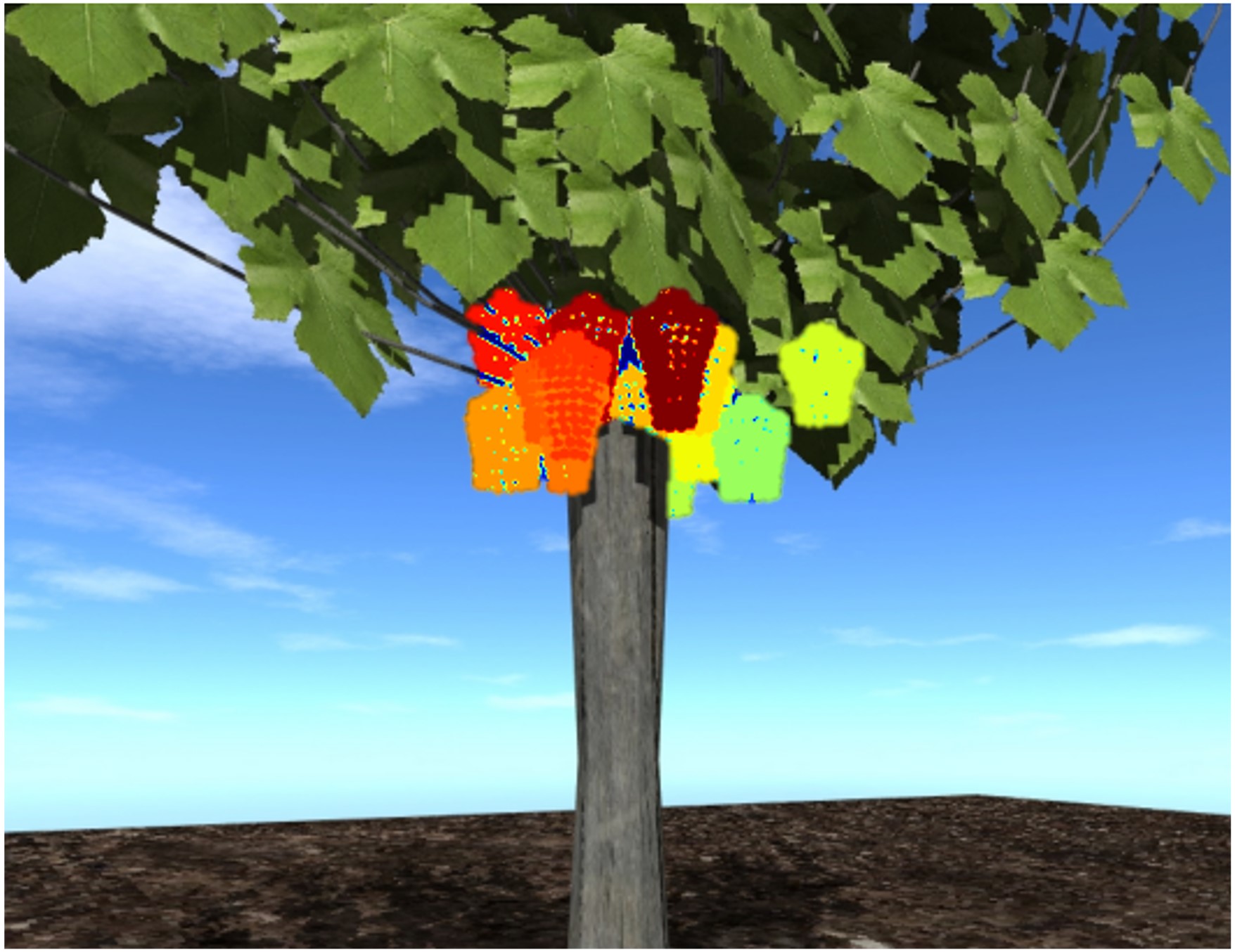}} \\
    \subfloat[]{\label{fig:single_strawberry}%
    \includegraphics[width=.47\linewidth]{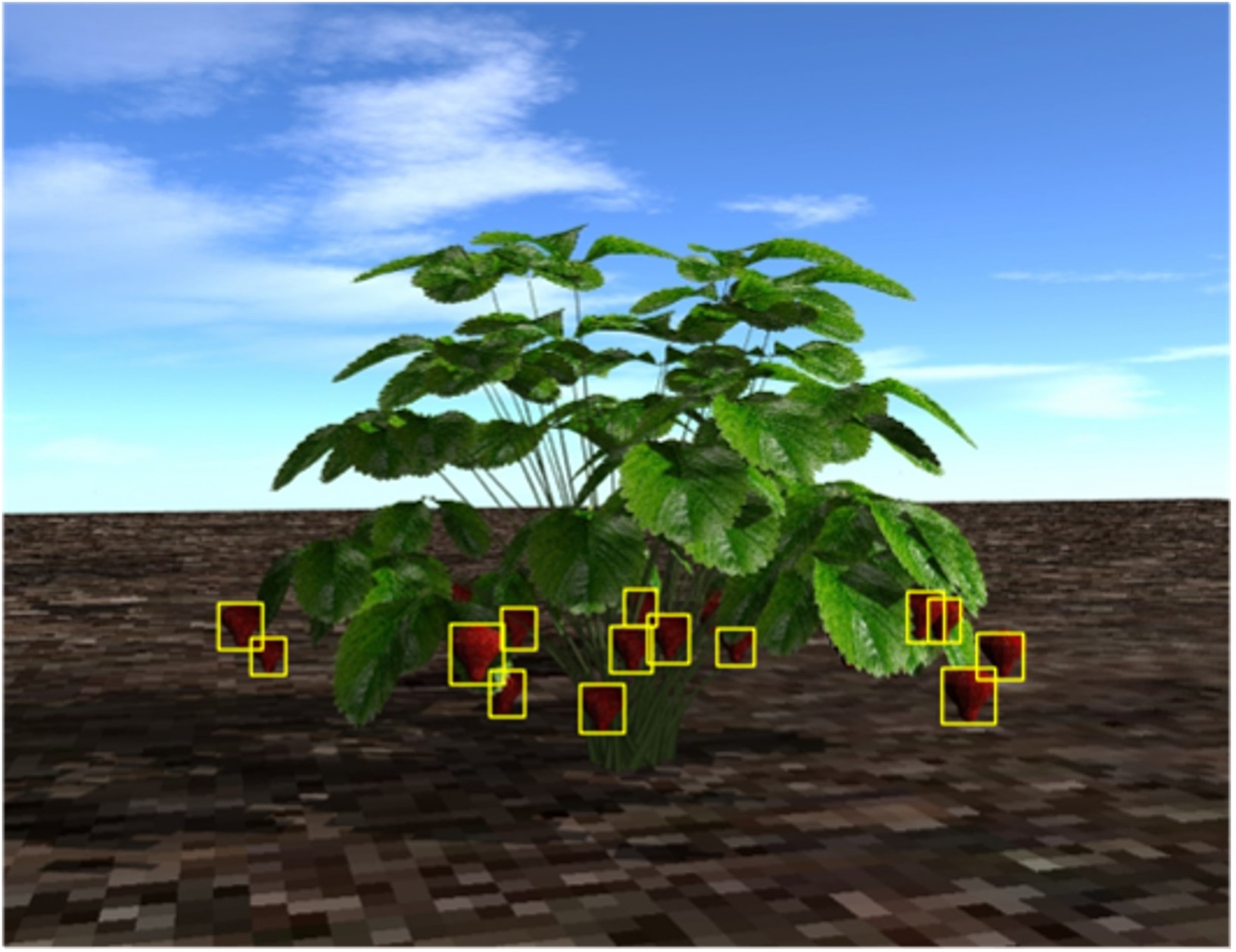}}
    \quad
    \subfloat[]{\label{fig:single_tomato}%
    \includegraphics[width=.47\linewidth]{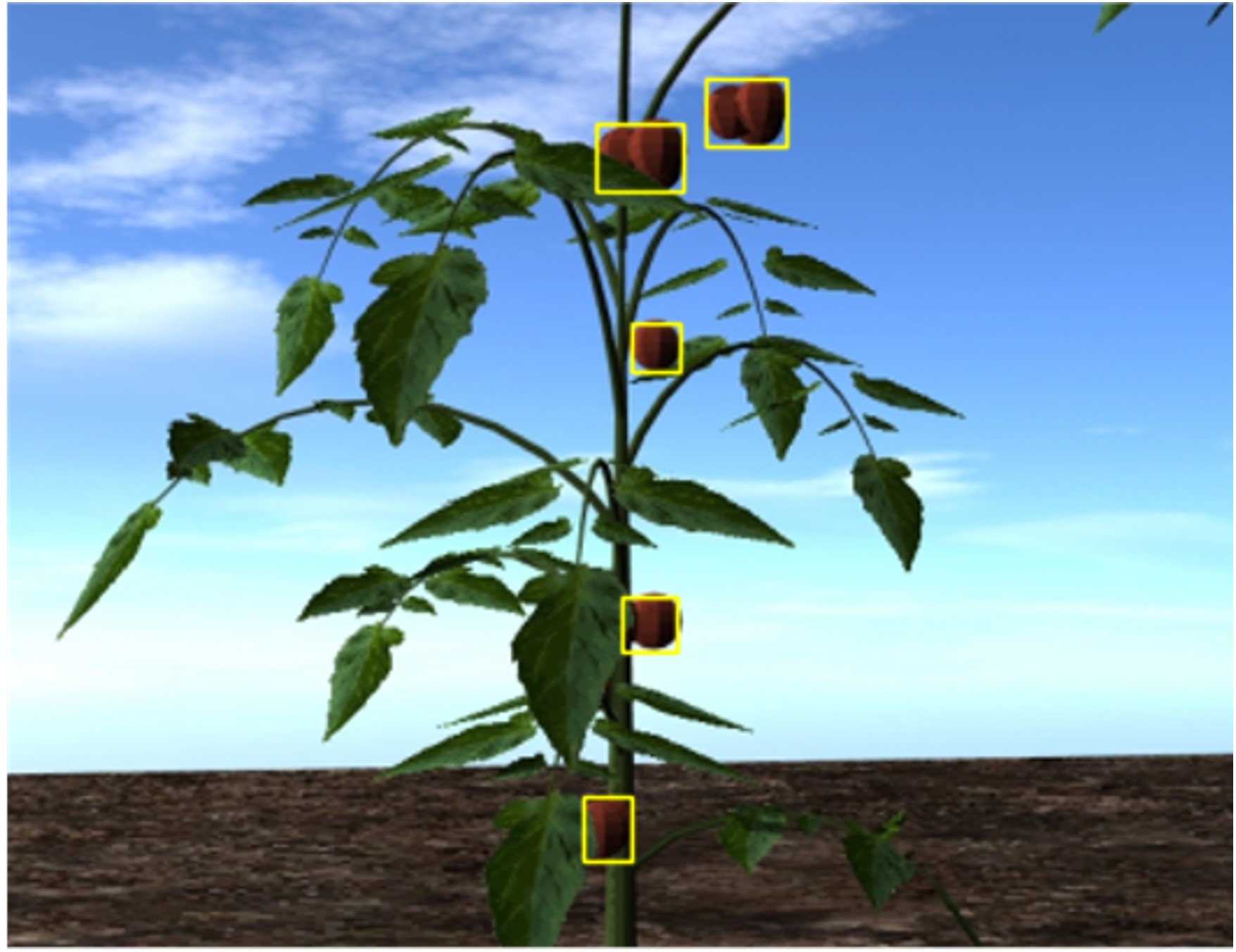}}
        \caption{
            Example images from single-plant scenarios of \mbox{DAVIS-Ag}:
            \protect\subref{fig:single_vine}--\protect\subref{fig:single_vine_seg}~goblet vine, 
            \protect\subref{fig:single_strawberry}~strawberry, and
            \protect\subref{fig:single_tomato}~tomato. 
            Labels for fruits are also visualized with 
            bounding boxes in \protect\subref{fig:single_strawberry}--\protect\subref{fig:single_tomato}, 
            and instance segmentation of \protect\subref{fig:single_vine} 
            in~\protect\subref{fig:single_vine_seg}.
            }
        \label{fig:single_plant_ex}
\end{figure}

For precision agriculture, accurate perception is an essential functionality for 
robots to identify the maturity or health statuses of plants.
One of the challenges is, however, caused by the wild environments, in which various 
objects such as fruits, stems, and leaves can be only partially visible due to 
occlusions by one another~\citep{KWBHV21}. 
For instance, a diseased fruit could be more easily misclassified as an normal instance 
when the anomalous parts are not fully visible~\citep{CWSC22}. 
In addition, yield estimation can be inaccurate, if a crop detector cannot access the views 
to some of occluded individuals~\citep{OSFSRBE22}.

As a potential solution, ``active vision'' approaches~\citep{B88} have recently been developed, 
in which an embodied agent plans its motion sequence to gain more informative viewpoints
around plants. 
More specifically, a robotic arm with some camera sensor on it is often deployed to move 
effectively while maximizing the area coverage of fruit to potentially classify 
its maturity~\citep{LTEM19,VHKE22}, predict the size and pose information~\citep{ZSMB21},
or reconstruct its complete $3$D structures~\citep{BVK22,ZZB22,MZB22}.
However, every model has been built up on a uniquely customized environment within 
either simulation or physical setups that could not be easily replicated 
without technical familiarity when new studies try to benchmark their novel approaches. 

\begin{figure*}[t!]\centering
    \subfloat[Backward]{\label{fig:move1}%
    \includegraphics[width=.25\linewidth]{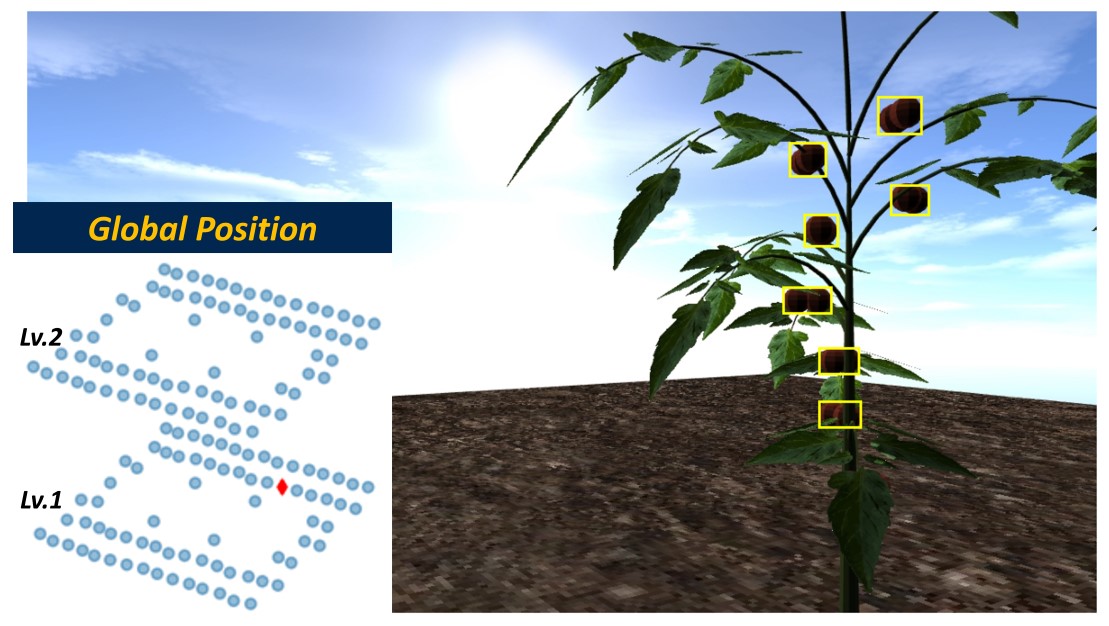}}
    \subfloat[Right]{\label{fig:move2}%
    \includegraphics[width=.25\linewidth]{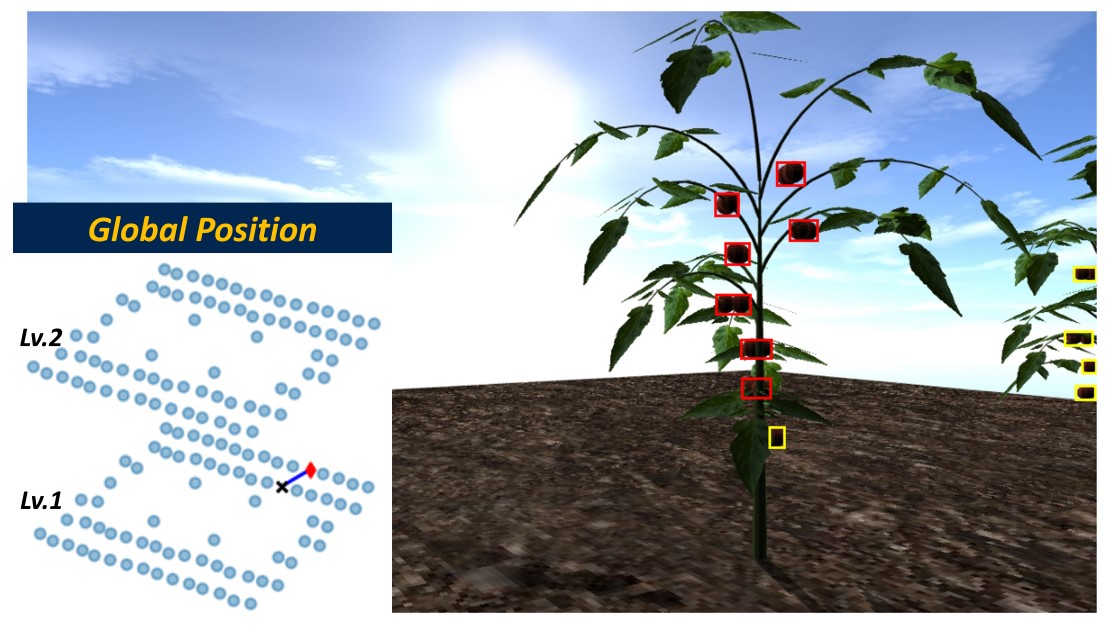}}
    \subfloat[Rotate CCW]{\label{fig:move3}%
    \includegraphics[width=.25\linewidth]{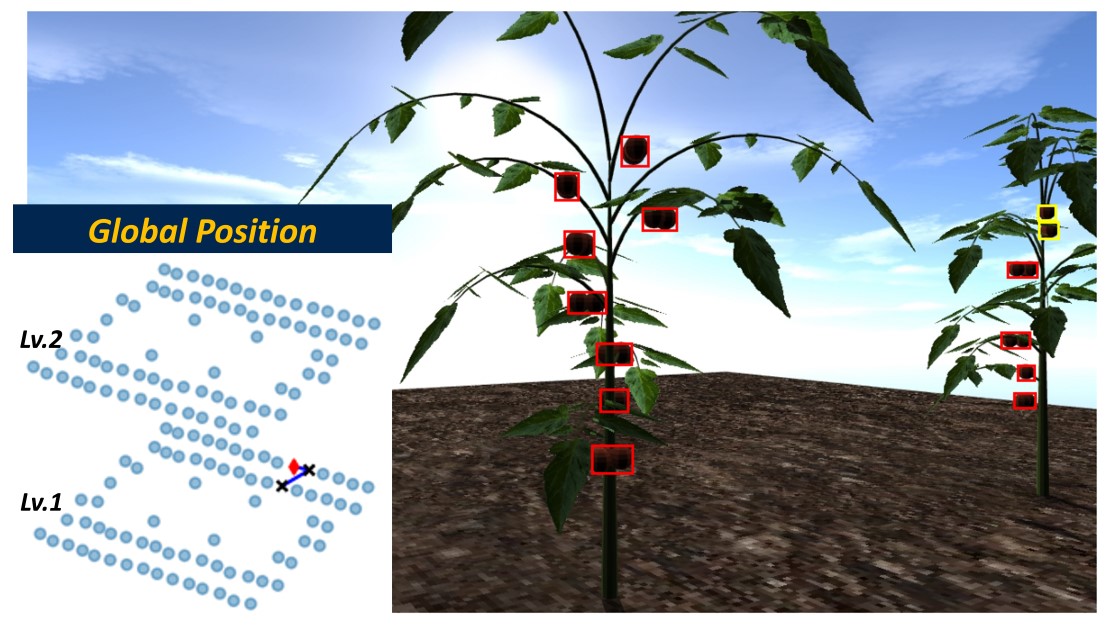}} 
    \subfloat[Forward]{\label{fig:move4}%
    \includegraphics[width=.25\linewidth]{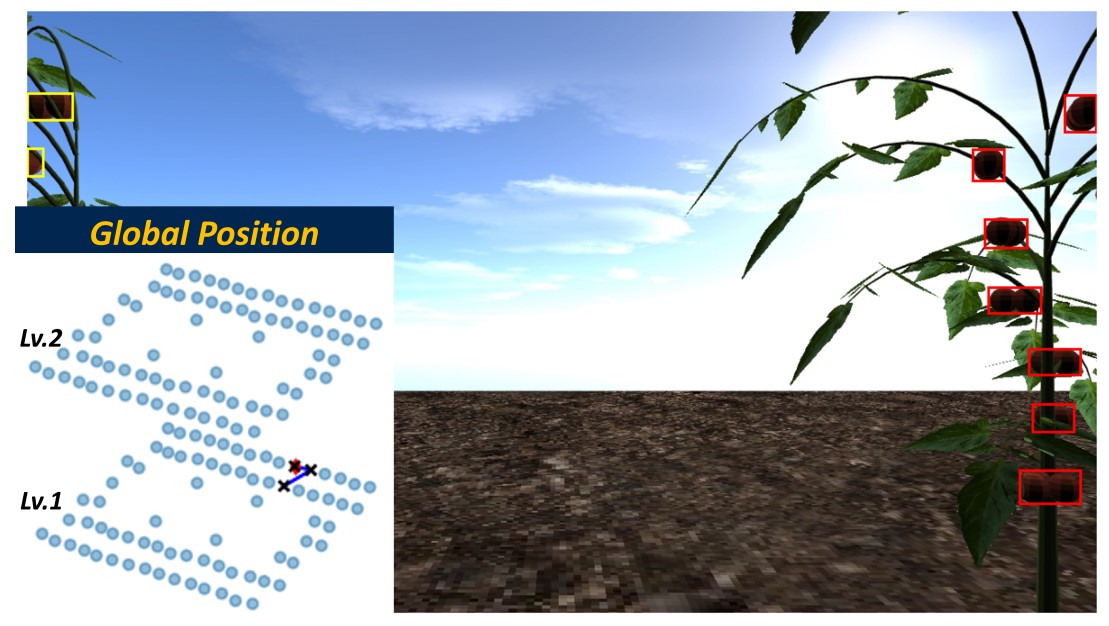}} \\
    \subfloat[Rotate CW]{\label{fig:move5}%
    \includegraphics[width=.25\linewidth]{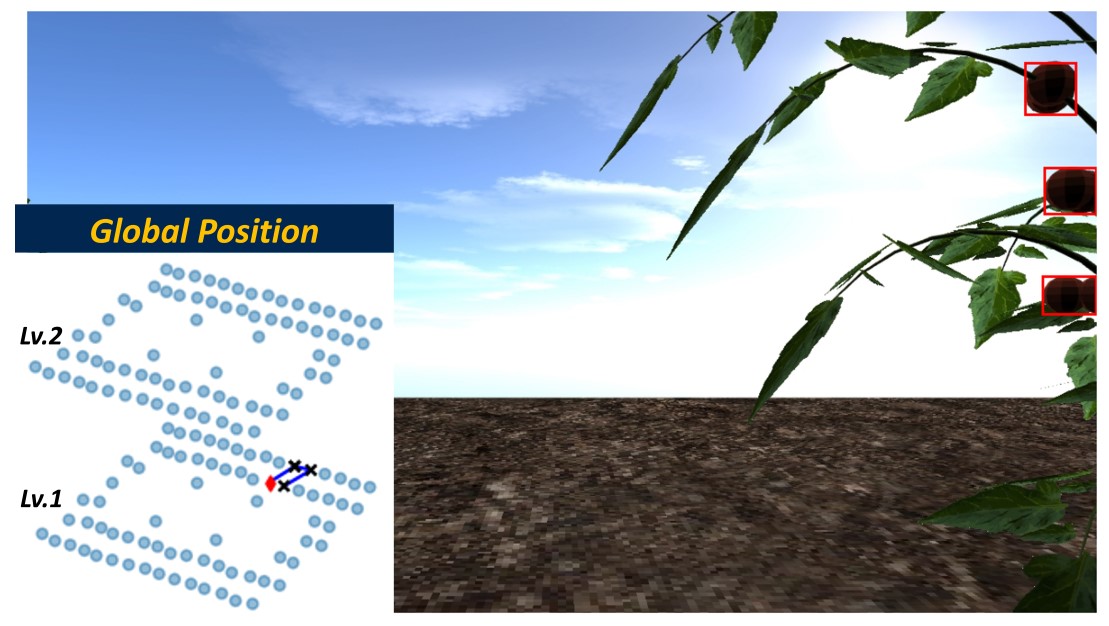}} 
    \subfloat[Up]{\label{fig:move6}%
    \includegraphics[width=.25\linewidth]{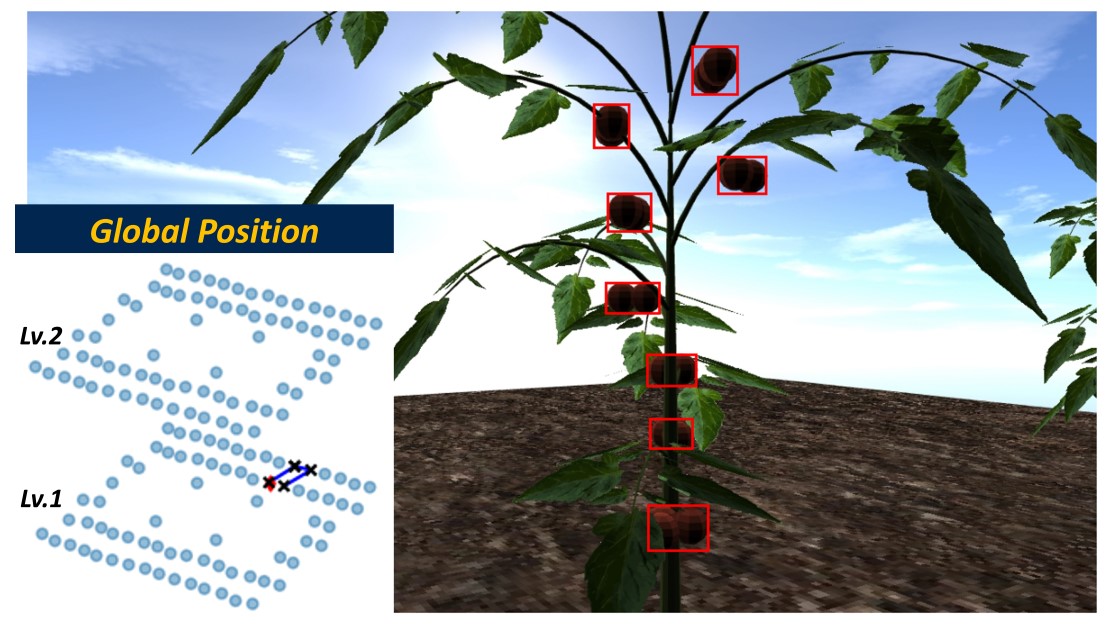}} 
    \subfloat[]{\label{fig:move7}%
    \includegraphics[width=.25\linewidth]{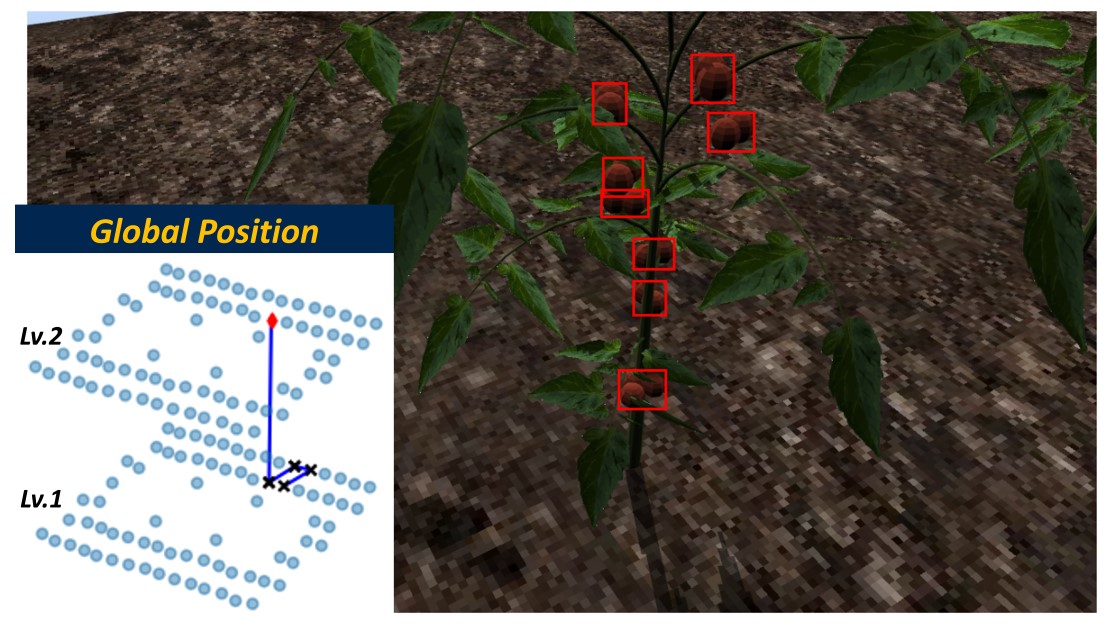}} \\
    \subfloat[]{\label{fig:fruit_size}%
    \includegraphics[width=.31\linewidth]{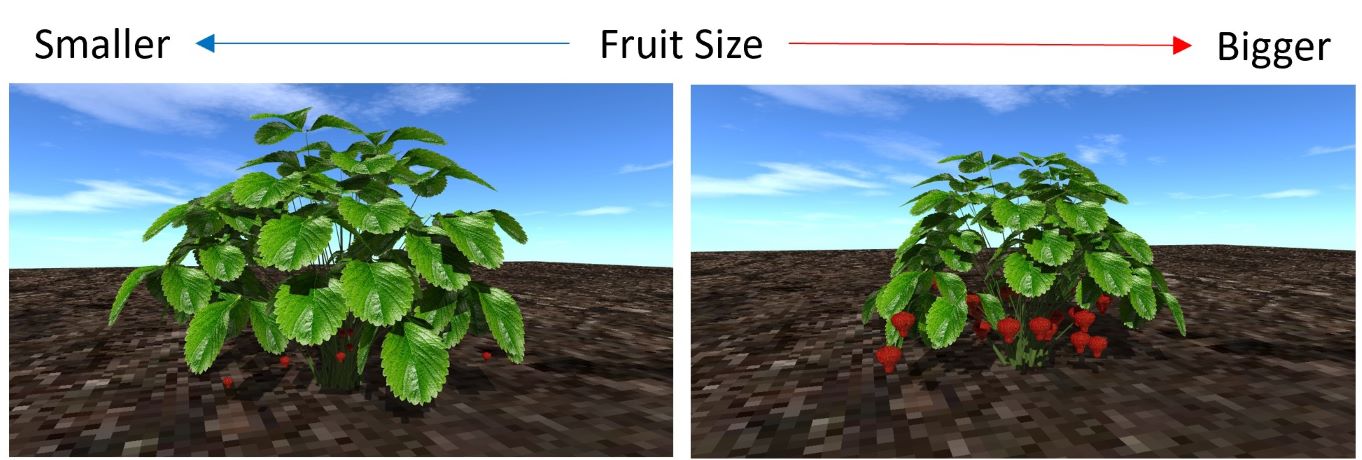}} 
    \quad
    \subfloat[]{\label{fig:leaf_length}%
    \includegraphics[width=.31\linewidth]{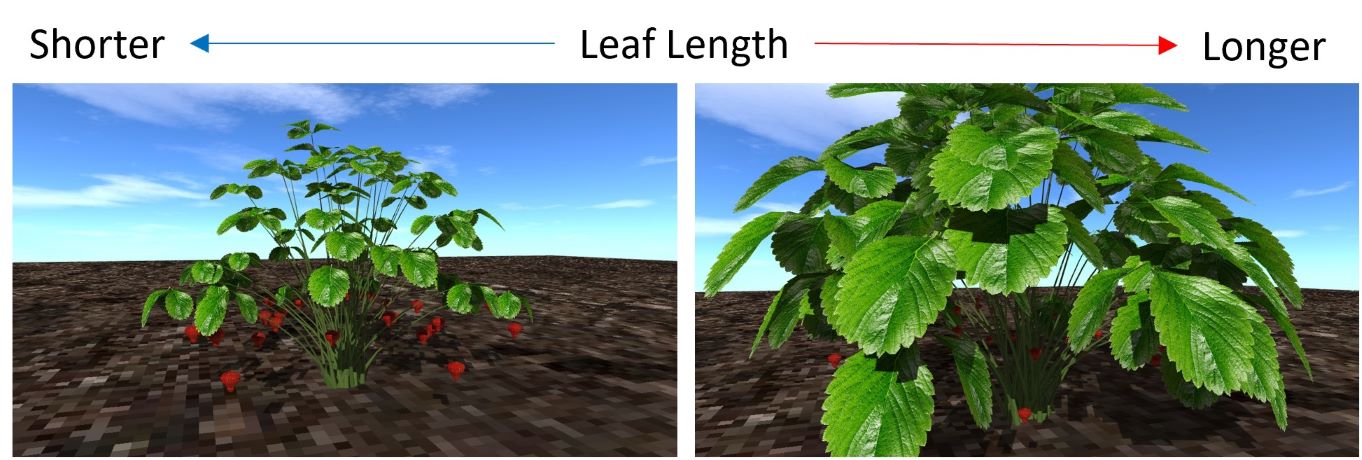}} 
    \quad
    \subfloat[]{\label{fig:plant_height}%
    \includegraphics[width=.31\linewidth]{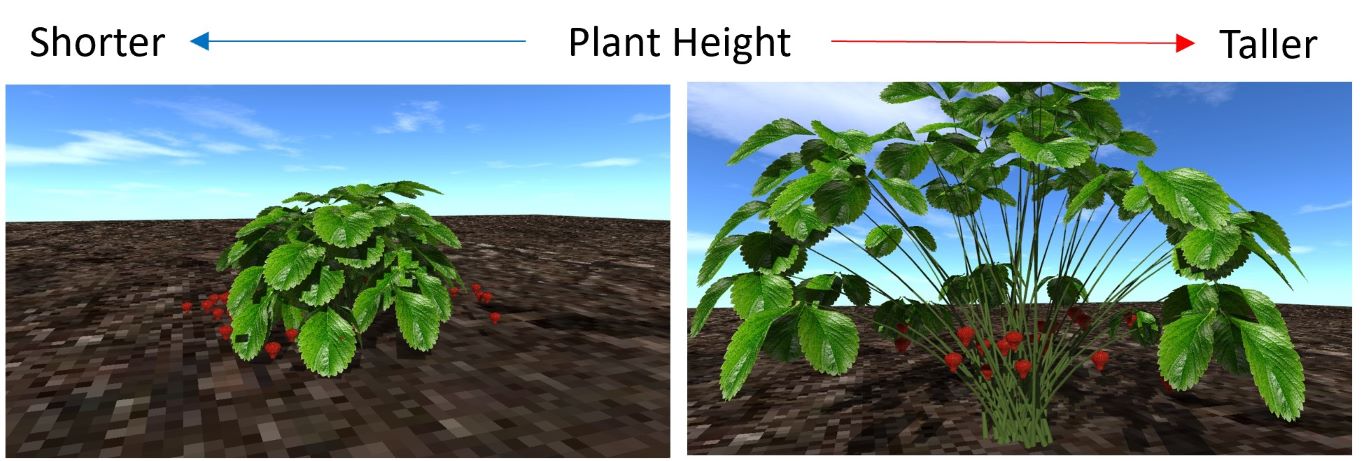}} \\
        \caption{
            \protect\subref{fig:move1}--\protect\subref{fig:move7}: Example trajectory from $t=0$ to $t=6$, applying six actions in a Multi-Plant Tomato scene. For each RGB~image, the caption indicates the action to be taken, and bounding boxes around tomatoes and global locations are also visualized for readers. For instance, the yellow (red) boxes surround unseen (seen) fruits, and the global map shows current positions with red diamonds and the traversed path as the solid line. \protect\subref{fig:fruit_size}--\protect\subref{fig:plant_height}: Single strawberry environments with various phenotypic characteristics intentionally generated to increase diversity in the dataset.}
        \label{fig:ex_trajs}
\end{figure*}

To bridge this gap, in this paper, we introduce an easy-to-access dataset, so-called \textbf{D}omain-inspired \textbf{A}ctive \textbf{VIS}ion in \textbf{Ag}riculture~(DAVIS-Ag), which contains over $502$K~HD-quality RGB~images gathered from $30$K~sampling  locations with diverse viewpoints around realistically simulated plants under varying phenotypic parameters (cf.~\autoref{fig:single_plant_ex} and \autoref{fig:fruit_size}--\ref{fig:plant_height}).
Specifically, $634$~synthetic orchards of strawberries, tomatoes, and grapes are considered at two different scales (i.e.,~Single-Plant and Multi-Plant).
Moreover, unlike typical agricultural image datasets~\citep{HRI20}, \mbox{DAVIS-Ag} provides the ``pointers'' between reachable viewpoints by possible actions 
(i.e.,~\emph{forward, backward, left, right, rotate, up, down}, etc.) to simulate sequences of active motions of embodied agent in orchards (cf.~\autoref{fig:move1}--\ref{fig:move7}).
Visual labels---i.e.,~bounding box and instance 
segmentation---of every observable fruit (\autoref{fig:single_plant_ex}) are also included for research in agricultural active vision. 


To the best of our knowledge, \mbox{DAVIS-Ag} is the \emph{first} public dataset 
purposed to assist in active vision research particularly in agricultural domains.
In addition, we present motivating examples to quantify and visualize the impact of occlusions on the levels of fruit visibility in plant environments. 
Furthermore, we provide experimental results of several baseline models as benchmarks on the task of target visibility maximization, using this dataset. 
Additionally, transferability to real strawberry environments is
tested to examine both visual fidelity and utility for prototyping in real-world applications.
To promote future research, our dataset is made publicly available
online. 

\begin{figure}[t]\centering
    \includegraphics[width=1.\linewidth]{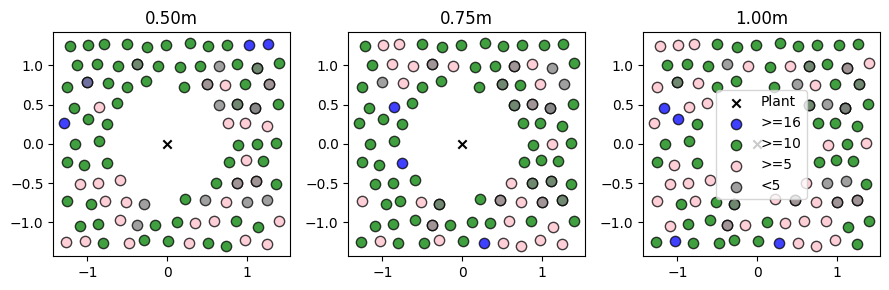}
    \caption{
        Colors represent the numbers of visible strawberries from different positions, visualized by circles, in an example environment where a single plant at the center has $24$~fruits in total. Heights of $0.50$m, $0.75$m, and $1.00$m were considered. 
        }
    \label{fig:motiv_spatial}
\end{figure}

\section{RELATED WORK}
\label{sec:related_work}


\subsection{Active Vision for Non-Agricultural Applications}
\label{sec:av_general}

Active vision has been broadly studied for decades in robotics~\citep{B88}, and 
it includes subfields of various tasks, such as classification~\citep{YRXCCPB19, SPCN21}, 
detection~\citep{FLWXL22}, and manipulation~\citep{CAF18} of objects, 
segmentation~\citep{NPGS21} and reconstruction~\citep{ZLRLGLHY22} of scenes, 
and also searching for certain items in environments~\citep{YLLZY18, SLF19}. 
Though objective functions can differ depending on the scenario, the same 
goal is essentially pursued to optimize the next poses of agent with visual sensors
to gain the most useful information for vision tasks. 


\subsubsection{Table-top Datasets}
\label{sec:table_top_data}

T-LESS~\citep{HHOMLZ17}, BigBIRD~\citep{SSNAA14}, and ROBI~\citep{YGLW21} each provide RGB~images along with depth data for a single or multiple objects arbitrarily posed on a ``turntable'' over which cameras were systematically moved along a spherical grid to emulate possible motions of an robotic arm. 
Single-plant scenarios in \mbox{DAVIS-Ag} are similarly designed to simulate a mobile robot, moving around a target plant located at the center of the field.

\subsubsection{Scene Understanding}
\label{sec:scene_datasets}

Similar to \mbox{DAVIS-Ag}, Active Vision Dataset (AVD)~\citep{APPKB17} and 
Real $3$D Embodied Dataset (R$3$ED)~\citep{ZZWQL22} both offer not only 
bounding boxes of objects in each image but 
the links between densely sampled viewpoints if they are reachable by an 
application of one of six actions---i.e.,~\texttt{forward}, \texttt{backward}, 
\texttt{left}, \texttt{right}, \texttt{rotate{ }clockwise}, and 
\texttt{rotate{ }counterclockwise}---to simulate an exploratory robot in a scene.
Yet, ours includes ``vertical'' translations such as \texttt{up} and \texttt{down} 
as well to emulate freer motions in agricultural open fields, which might be particularly useful to avoid occluders 
through highly complex structures of plant.  
We also add some level of white noise to the camera positions to account for 
possible ``slips'' of robot in unfavorable terrain conditions, while AVD~\citep{APPKB17} and 
R$3$ED~\citep{ZZWQL22} do not incorporate this into their indoor navigation. 

More recently, photorealistic $3$D~simulators e.g.,~Habitat~$3.0$~\citep{AIHabitat3}) have been introduced to enable virtual embodied agents to interact with 
physics-engine-based environments which also offer realistic visualizations. 
As in~\citep{DMDQZMS23}, however, \mbox{DAVIS-Ag} is a pre-computed dataset that includes all possible visuals for spatial exploration, enabling researchers to incorporate them easily into 
their framework namely on any development environment.
Furthermore, while all existing platforms mentioned above were created to visualize indoor environments,
\mbox{DAVIS-Ag} is uniquely designed for exploration in agricultural ``outdoor'' fields with distinct objects, such as fruits, leaves, and stems. 

\begin{table}[]\centering
    \small
    \begin{tabular}{|c||c|c|c|}
    \hline
                      & Strawberry & Tomato & Goblet Vine   \\ \hline\hline
     Plant/Trunk Height        & $40$    & $100$ & $70$  \\ \hline
     Fruit Radius     & $2.5$ & $3$ & $0.75$  \\ \hline
     Leaf Length/Width & $10$  & $20$ & $18$  \\ \hline
     Camera Altitudes & $25, 40, 55$ & $70, 110, 150$ & $50, 75, 100$ \\ \hline
    \end{tabular}
    \caption{Top three rows show the default values of parameters for plant generation in Helios, 
            in which the fruit radius of vine refers to that of individual 
            berries.
            The last row reveals the three camera heights considered for each orchard 
            type. Every value is in units of centimeters.             
            }
    \label{tab:default_plant_params}
\end{table}

\subsection{Active Vision in Agriculture}
\label{sec:av_ag}

In agriculture, active vision has been investigated to better monitor statuses of
plants, localize fruits for robotic picking, or predict potential yields under 
significantly cluttered environments.  
For example, dynamic viewpoint-decision systems were developed to strategically 
move a robotic arm to accurately estimate the maturity~\citep{VHKE22} or the 
size~\citep{ZZB22} of partially visible fruit. 
In addition, maximizing the viewed area of target fruit has been studied to 
closely examine the instance as well as provide a manipulator with useful 
information for picking~\citep{ZZB22, LTEM19}.
For similar motivation, the authors in~\citep{MZB22} used active motions of a robotic arm with 
a sensor to reconstruct the hidden structure of fruit under another object. 
Full $3$D reconstruction of entire plant also has been performed 
in the context of active vision~\citep{BVK22,GPFWMP19}.

Nonetheless, absence of a common testbed---such as ones discussed 
in~\autoref{sec:av_general}---has been a 
challenge to evaluate a method against another in agricultural applications. 
For instance, most of the employed simulators have been implemented 
based upon V-REP~\citep{RSF13}, Panda$3$D~\citep{GM04}, or 
Robot Operating System~(ROS)~\citep{QCGFFLWN09} with Gazebo~\citep{KH04} which 
are specially customized for each research problem. 
In this paper, on the other hand, we propose \mbox{DAVIS-Ag}, which does not require 
technical familiarity with any of those for simulating robot motions, since ``view-to-view'' pointers, linking reachable view images, are available. 

In a similar sense, \mbox{DAVIS-Ag} is distinct from other agricultural datasets, which only provide the images that were gathered independently for traditional computer vision tasks~\citep{HRI20, CLSWBS17, WMMCSRCSB23}. 
It is also different from the datasets for localization tasks~\citep{MTABLDT23, PMHPTGLTCH23}---generally collected using camera sensors, inertial measurement units, and wheel encoders as mobile robots travel along fixed paths---since their sparse samples do not allow for simulations of novel trajectories.

\begin{figure*}[t!]\centering
    \subfloat[]{\label{fig:motiv_diff}%
    \includegraphics[width=.48\linewidth]{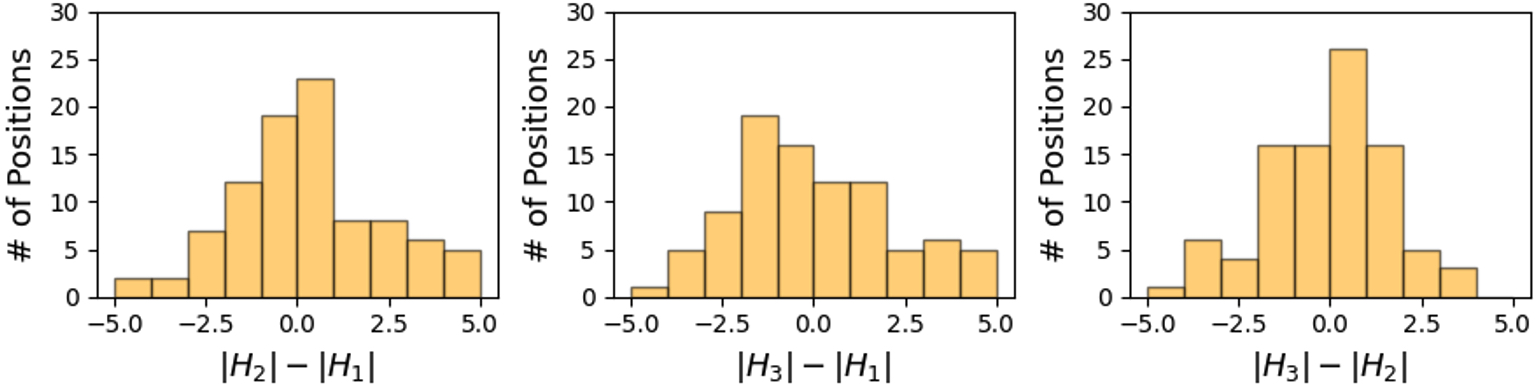}}
    \subfloat[]{\label{fig:motiv_novel}%
    \includegraphics[width=.48\linewidth]{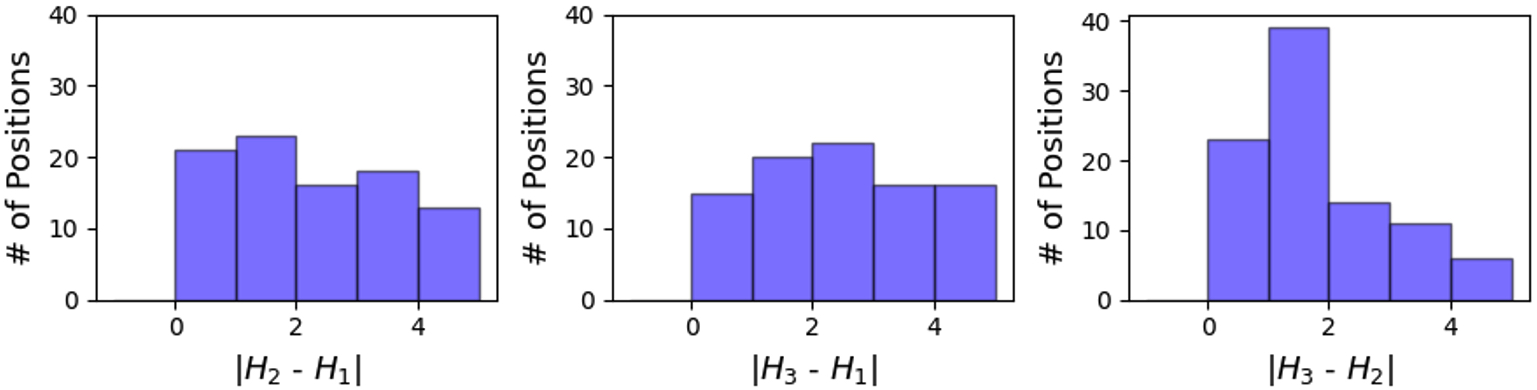}}
        \caption{            
            Histograms depicting \protect\subref{fig:motiv_diff}~the differences in the numbers of strawberries observed at various altitudes 
            and \protect\subref{fig:motiv_novel}~the numbers of 
            newly found instances at higher altitudes. 
            $H_k$ denotes the set of observable fruits at a particular altitude~$z_k$ at fixed locations within the example of~\autoref{fig:motiv_spatial}, where $z_1 < z_2 < z_3$.
            }
        \label{fig:motiv_diff_novel}
\end{figure*}

\section{DATA SYNTHESIS \& COLLECTION}
\label{sec:data_collection}

This section describes more technical details on our pipeline for simulation and procedure of data collection. 

\subsection{$3$D Plant Simulation}
\label{sec:3d_plant_sim}

Our data synthesis pipeline is built based upon our simulation toolbox~\citep{GJRFBE23}, leveraging the AgML\footnote{https://github.com/Project-AgML/AgML}
and the Helios~\citep{B19} frameworks.
Specifically, AgML is utilized to interface with the Helios plugins, visualizing simulated three-dimensional plants and environments through OpenGL~\citep{B19}. 
In particular, these simulation outcomes encompass realistic, diverse shapes of plants generated based on semi-random tree geometries~\cite{weber1995}. 
Specifically, we configured AgML to utilize several useful plugins of Helios as below: 
\begin{itemize}
    \item \emph{Canopy Generator}: Synthesized three kinds of crops, including \texttt{goblet vines}, \texttt{strawberries}, and \texttt{tomatoes} (cf.~\autoref{fig:single_plant_ex}), varying their sizes of fruit, leaf, and trunk within the ranges of $\pm20\%$, $\pm20\%$, and $\pm15\%$ from the default settings in~\autoref{tab:default_plant_params}. 
    \item \emph{Visualizer}: Generated RGB images of $1280\times720$ in resolution from a number of simulated camera positions densely distributed across the field (cf.~\autoref{sec:scenarios_camera_move}).
    \item \emph{Synthetic Annotation}: Produced fruit-related labels for each image (cf.~\autoref{fig:single_vine_seg}--\ref{fig:single_tomato}): (1)~$2$D coordinates of bounding boxes  and (2)~instance segmentation masks. 
\end{itemize}


\begin{table}[]\centering
    \small
    \begin{tabular}{|C{10mm}||C{10mm}|C{10mm}|C{10mm}|C{10mm}|C{12mm}|}
    \hline
      & Step Size & Heights & Angles  & Actions & Images \\ \hline\hline
     SP & $25$cm & $3$ & $1$ & $6$ & \footnotesize$285\sim348$ \\ \hline
     MP  & $50$cm & $2$ & $12$ & $8$ & \footnotesize$1,128\sim1,800$ \\ \hline
    \end{tabular}
    \caption{Left four columns show the settings for SP and MP~scenarios, 
    including the numbers of possible camera altitudes and angles, while
    the right-most column indicates the number of images per virtual farm in 
    each scenario.}
    \label{tab:single_multi}
\end{table}

\subsection{Two-Scale Scenarios \& Camera Poses}
\label{sec:scenarios_camera_move}

For each kind of plant, we particularly consider two types of scenarios---i.e.,~``Single-Plant''~(SP) and ``Multi-Plant''~(MP)---with the camera settings in~\autoref{tab:single_multi}.
Specifically, in SP, each camera view is set up to always aim at the plant at the origin regardless of its spatial 
position, with a reasonable assumption that tracking a single plant target to maintain 
it within the view can be easily implemented. 
The set of possible actions includes \emph{forward, backward, left,} and \emph{right} along with 
\emph{up} and \emph{down}, which control the altitude of viewpoint between three levels specified in~\autoref{tab:default_plant_params}. 

On the other hand, MP~uses three plants in a single row for both tomato and goblet vine cases albeit five in a row for strawberry 
fields.
Additionally, the RGB~images are sampled every $30$~degrees at each position as 
in~\citep{APPKB17}, and as a result, \emph{rotate\_clockwise} and \emph{rotate\_counterclockwise}
are added to the action set (cf.~\autoref{fig:move1}--\ref{fig:move7}). 
To compensate for this increased computational load in data collection, we set the sampling 
resolution to be slightly coarser by doubling the step size from~$25$cm~\citep{ZZWQL22} to~$50$cm and 
considering only two levels of camera altitudes---i.e.,~the highest and the lowest. 

As in~\autoref{fig:motiv_spatial}, contrary to existing datasets~\citep{APPKB17,ZZWQL22}, 
we use additive white Gaussian noise with a standard deviation of $2.5$cm on each 
$x,y$-coordinate of the camera position to simulate a robot ``slipping'' in harsh terrain conditions of real outdoor environments.
Moreover, the spatial sizes of grid in the two scenarios are set up differently---i.e.,~$3$m $\times$ $3$m for SP and $7$m $\times$ $3$m for MP---to keep the space to be sufficiently large for exploration while achieving reasonable computational costs for simulation. 

\begin{table}[]\centering
    \small
    \begin{tabular}{|C{10mm}||c|c|c|c|c|c|}
    \hline
      & \multicolumn{2}{c|}{Strawberry} & \multicolumn{2}{c|}{Tomato} & \multicolumn{2}{c|}{Goblet Vine} \\ \hline
       & SP & MP & SP & MP & SP & MP   \\ \hline\hline
     Scenes & \scriptsize$86$ & \scriptsize$77$ & \scriptsize$130$ & \scriptsize$113$ & \scriptsize$182$ &  \scriptsize$44$ \\ \hline
     Images & \scriptsize$24,510$ & \scriptsize$86,856$ & \scriptsize$45,240$ & \scriptsize$203,400$ & \scriptsize$63,336$  & \scriptsize$79,200$  \\ \hline
        \end{tabular}
    \caption{Data size for each type of plant in \mbox{DAVIS-Ag}.}
    \label{tab:data_spec}
\end{table}

\subsection{Labels \& Potential Applications}
\label{sec:annotations}

The following labels are available for each RGB~image:

\begin{itemize}
    \item \textbf{2D bounding boxes} \& \textbf{ID's} for individual fruits (\autoref{fig:single_tomato}).
    \item \textbf{Instance segmentation mask} for each fruit (\autoref{fig:single_vine_seg}).
    \item \textbf{Pointers} to other images reachable by an action.
    \item \textbf{Camera pose}, denoted as $(x,y,z, \psi, \theta)$ on the global coordinate system, where 
            $(x,y,z)$ represents the position in three-dimensional space while $\psi$ and $\theta$ are the yaw and pitch of camera, respectively. 
\end{itemize}

Bounding boxes around observed fruits are generally useful for building fruit detectors
for yield estimation~\citep{OSFSRBE22}, robotic picking, or health monitoring. 
In particular, identifying individual crops~\citep{KMC21} could also be developed from the labels of instance ID. 
Similarly, instance segmentation could also be useful to find the specific examples of occluded fruit so that a robot could learn an active maneuver to lead better views for ``fruit coverage''~\citep{ZLMB21, LTEM19}, which infers poses or sizes of individual fruits.
In addition, 
the global camera poses may be useful when validating frameworks for ``visual localization''~\citep{PMHCH22} in agricultural environments.
``Vision-based navigation'' could also be investigated with this dataset, building robots to successfully arrive in a queried view, while only relying on camera sensors. 
Note that all these domains requiring robot motions can be considered due to the availability of the ``view-to-view'' pointers. 

For post-processing, we discarded unreasonably small bounding boxes or ones that barely show a fruit by employing a threshold approach.
The minimum numbers of pixels for single strawberries, tomatoes, and grapes were set to $210$, $240$, and $700$, respectively. 
Additionally, if an action leads to a location either outside the grid or overly close to a plant (e.g.,~$<75$cm for strawberry and $<100$cm for others), it is regarded 
as an invalid action, connecting to no image.

\subsection{Data Specification}

We finally produced $>502$K~images from over $632$~virtual farms to construct the \mbox{DAVIS-Ag} dataset. 
Consequently, our dataset is magnitudes larger than other non-agricultural active vision datasets such as AVD~\citep{APPKB17} and R$3$ED~\citep{ZZWQL22}.
Additional specifications are presented in~\autoref{tab:data_spec} for further details. 

\begin{table}[]\centering
    \small
    \begin{tabular}{|c|c||C{10mm}|C{10mm}|C{10mm}||C{10mm}|}
    \hline
      Scenario & Metric & LRD & UDD & Rand & A2C \\ \hline\hline
      \multirow{3}{*}{SP-ST} & $OER_\tau$ & $.433$ \scriptsize$\pm.007$ & $.371$ \scriptsize$\pm.008$ & $.438$ \scriptsize$\pm.101$ & $\textbf{.520}$ \scriptsize$\pm.004$ \\ 
                           & \small Cum.~Reward & $.026$ \scriptsize$\pm.002$ & $.008$ \scriptsize$\pm.002$ & $.027$ \scriptsize$\pm.002$ & $\textbf{.052}$ \scriptsize$\pm.001$ \\ 
                           & \small Ep.~Length & $2.157$ \scriptsize$\pm.043$ & $2.137$ \scriptsize$\pm.036$ & $2.652$ \scriptsize$\pm.056$ & $\textbf{2.987}$ \scriptsize$\pm.008$ \\ \hline
    \multirow{3}{*}{MP-GV} & $OER_\tau$ & $.316$ \scriptsize$\pm.008$ & $.313$ \scriptsize$\pm.006$ & $.340$ \scriptsize$\pm.002$ & $\textbf{.369}$ \scriptsize$\pm.009$ \\ 
                           & \small Cum.~Reward & $.006$ \scriptsize$\pm.002$ & $.004$ \scriptsize$\pm.001$ & $.012$ \scriptsize$\pm.002$ & $\textbf{.020}$ \scriptsize$\pm.003$ \\ 
                           & \small Ep.~Length & $2.558$ \scriptsize$\pm.097$ & $2.683$ \scriptsize$\pm.118$ & $4.163$ \scriptsize$\pm.058$ & $\textbf{5.000}$ \scriptsize$\pm.000$   \\ \hline
        \end{tabular}
    \caption{Results from evaluated methods on SP-Strawberry~(SP-ST) and MP-Goblet Vine~(MP-GV) test data. Each average score and standard deviation was computed from three sets of $200$~random episodes.} 
    \label{tab:sp_mp_result}
\end{table}

\section{MOTIVATING EXAMPLE}
\label{sec:motivating_ex}

Here, we use a representative instance of simulated plant environment to showcase that  
active viewpoint planning can considerably impact performances 
in agricultural tasks that would particularly require clear visibility to a certain 
type of object (e.g.,~fruit). 
Specifically, as visualized in~\autoref{fig:motiv_spatial}, we examined the numbers of visible 
strawberries from different viewpoints in a field where a single plant with 
$24$~fruits were located at the center of a grid. 
All configurations for the poses of deployed camera and the environmental settings in 
this \emph{single-plant} scenario followed the details in~\autoref{sec:scenarios_camera_move}. 

\Autoref{fig:motiv_spatial} visualizes the dramatic difference in 
visibility of fruits depending on the camera location; for example,  
the number of seeable fruits can significantly vary from $<5$ to $\geq16$ simply 
by repositioning the camera only several steps away. 
In addition, none of the spots provided the viewpoint that can cover all of $24$~strawberries at once.
Similarly, \autoref{fig:motiv_diff} proves that even at the same location, using a higher or lower view can also lead to a change to the visibility.

These observations imply that 1)~plant environments are essentially highly 
unstructured, as demonstrated even with this simplified scenario, and thus, 2)~the application of 
\emph{active vision} can greatly benefit autonomous systems for comprehensive monitoring of whole plants or fruits.
The latter is particularly advocated by~\autoref{fig:motiv_diff_novel}, which shows 
that though lower views tend to show more strawberries 
(\autoref{fig:motiv_diff}), higher ones can still lead to 
the encounters with novel instances (\autoref{fig:motiv_novel}).
Hence, strategically choosing viewpoints needs to be considered to 
precisely understand an agricultural environment from perceived information. 
Our proposed dataset is designed to further encourage 
relevant research by providing a large amount of realistic plant images with visual and spatial annotations. 

\begin{figure}[t]\centering
    \subfloat[]{\label{fig:learn_curve}%
    \includegraphics[width=.42\linewidth]{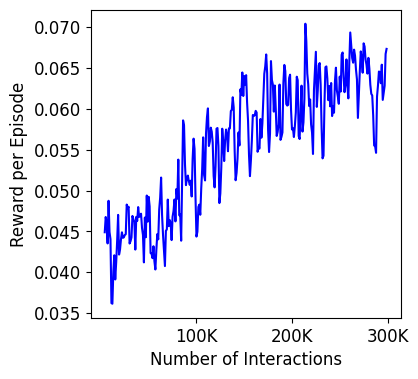}} 
    \quad
    \subfloat[]{\label{fig:oer_dist}%
    \includegraphics[width=.42\linewidth]{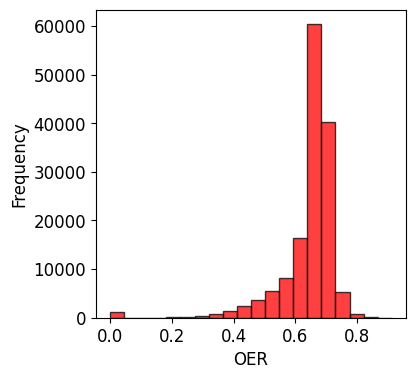}}\\
    \subfloat[]{\label{fig:sp_st_acts_dist}%
    \includegraphics[width=.40\linewidth]{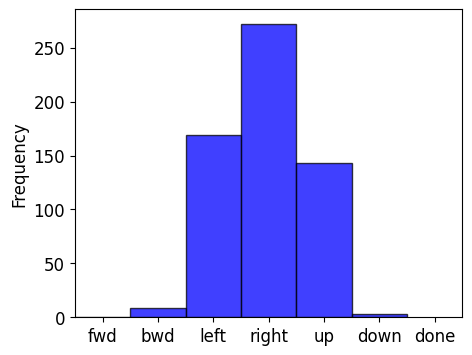}}
        \caption{
            \protect\subref{fig:learn_curve}~Episodic rewards over $300$K interactions while A$2$C learns the TVM~task from SP-Strawberry data, with some level of rolling average applied for clarity.
            Frequencies of \protect\subref{fig:oer_dist}~OER~scores for all available strawberry instances and             \protect\subref{fig:sp_st_acts_dist}~selected actions from tests. 
            }
        \label{fig:oer_dist_learn_curve}
\end{figure}

\section{EXPERIMENTAL SETTINGS}
\label{sec:experiments}
%


\subsection{Target Visibility Maximization}
\label{sec:occlusion_minimization}

Inspired by target area maximization~(TAM)~\citep{LTEM19}, which aims to maximize the viewed area of a target object,
we define the target visibility maximization~(TVM) problem
as maximizing the ``ratio'' of the target fruit's area to the size of the bounding box at the \emph{last} timestep. 
Compared to TAM, TVM~is designed to avoid incentivizing simplistic motions, such as moving forward to the target, to simply enlarge the target area without reduction of obstruction level.
TVM~aligns well with sub-tasks intrinsic to robotic harvesting~\citep{KWBHV21} and crop monitoring~\citep{CWSC22}, where robots are not just required to count individual crops but to gain full observations of their surface.

As in other active perception research~\citep{APPKB17}, a target fruit is assumed to be partially visible within the robot's field of view and localized with a bounding box by a well-trained object detector at the initial timestep. 
Thus, our focus is on enabling an active agent to plan sequential actions for minimization of the obstruction over the target fruit within the bounding box. 
For evaluation, we employ the metric ``object exposure rate''~(OER) after $\tau$~active motions: 
\begin{equation}
     OER_{\tau} \triangleq \frac{n}{(x_2-x_1)(y_2-y_1)},
     \label{eq:search_score}
\end{equation}
where $(x_1, y_1)$ represents the upper-left $x,y$~coordinates, and $(x_2, y_2)$ the lower-right coordinates after $\tau$~steps. 
Here, $n$~denotes the number of pixels belonging to the target fruit within the bounding box. 
A higher OER~score implies a more unobstructed view of the target object.


\subsection{Heuristic Baselines}
\label{sec:benchmarked_methods}

We first consider several heuristic approaches--$LRD, UDD,$ and $Rand$---selecting a random action~$a_t$ from one of the following action sets~$A$ at each timestep~$t$:

\begin{itemize}
    \item $A_{LRD} = \{left, right, done \}$ 
    \item $A_{UDD} = \{up, down, done \}$ 
    \item $A_{Rand} = \{forward, backward\} \cup A_{LRD} \cup A_{UDD}$
\end{itemize}
where \emph{done} action is for the robot to decide to stay at the same location and terminate the episode.
In the MP~scenarios, $A_{Rand}$ also includes the \emph{rotate cw} and \emph{rotate ccw}. 

\subsection{RL Agent \& Implementation}

We also train an agent using the Advantage Actor Critic~(A$2$C) reinforcement learning~(RL) algorithm~\citep{MBMGLHSK16}. 
To be specific, at each time~$t$, the A$2$C~agent decides its action~$a_t$ by processing an input~$\mathcal{X}_t \in \mathbb{R}^{224\times224\times4}$, which is obtained by concatenating the downsized RGB~image~$\mathcal{I}_t \in \mathbb{R}^{224\times224\times3}$ from the current viewpoint and the target bounding box~$\mathcal{B}_t \in \mathbb{R}^{224\times224\times1}$. In particular, $\mathcal{B}_t$ is a zero matrix with the values of one inside the bounding box' area. 

This action selection process is iterated $\tau$~times for an episode, and the sequence of actions is then evaluated based on the final~OER. Depending on the quality of obtained views, a reward~$r$ is computed in the following manner: $r = | max (OER_\tau - OER_0, 0) |^2$, where $OER_t$ denotes $OER$ at time~$t$. That is, the actions leading to an increase in the OER are reinforced to cause them to be more likely executed in similar visual states.

The neural network architecture consists of three convolutional layers---i.e.,~$(8, 3), (4, 2),$ and $(3, 1)$, indicating (kernel size, stride)---followed by a fully-connected layer to produce $512$-dimensional feature vectors.
Additionally, an action network and a value network utilize separate fully-connected layers to generate actions and estimate the state's value from these features, respectively.

\begin{figure}[t]\centering
    \includegraphics[width=.65\linewidth]{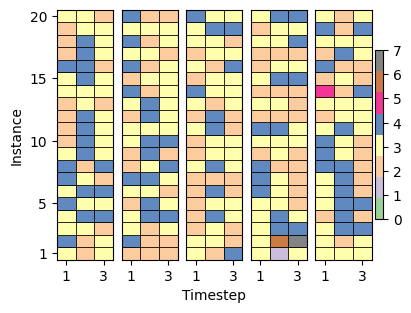}
        \caption{
            Action choices in $100$~representative test episodes in SP-Strawberry farms. Actions~$0$ to~$6$ correspond to \emph{forward, backward, left, right, up, down}, and \emph{done}, respectively.
            }
        \label{fig:sp_st_acts}
\end{figure}

\begin{table}[]\centering
    \small
    \begin{tabular}{|c|c|c|c|c|c|c|}
    \hline
     & \multicolumn{3}{|c|}{Real-Only} & \multicolumn{3}{|c|}{DAVIS-Ag $\rightarrow$ Real} \\ \hline
     Split & $1$ & $2$ & $3$ & $1$ & $2$ & $3$  \\ \hline
       mAP@$50$ &  $.943$ & $.888$ & $.920$ & \boldmath{$.945$} & \boldmath{$.911$} & \boldmath{$.936$} \\ \hline
       mAP@$50$:$95$ &  \boldmath{$.619$}  &  $.504$ &  $.505$ & $.618$ & \boldmath{$.511$} & \boldmath{$.517$} \\ \hline
        \end{tabular}
    \caption{Test performance for ripe strawberry detection under different splits employed out of~\citep{LBBP22}. The highest value for each split and each metric is in bold.} 
    \label{tab:detection_results}
\end{table}

\subsection{Task Configurations}
\label{sec:comparative_results}

We explore both SP and MP scenarios to test the active perception capabilities.  
Specifically, SP of strawberry data is employed for each episode with time steps~$\tau=3$, whereas MP involves the data of goblet vines with $\tau=5$. 
Each episode is terminated when either $\tau$~steps have been taken or \emph{done}~action is executed. 
Moreover, if the robot acts to move out of the global grid or to a location from which the target fruit is not viewed, that movement is ignored to keep the robot at the previous position. 
In experiments, the same split is applied to use $70\%$ and $30\%$~data for training and test environments, respectively, unless mentioned otherwise. 
Each episode is initialized with a random viewpoint showing a target fruit with an OER~$<.400$. 
\Autoref{fig:oer_dist} displays the right skewed distribution of OER's in the SP-Strawberry dataset, with $4\%$~instances satisfying this condition. 
Note that while each test session involves $200$~episodes, 
the average performance over three independent sessions is reported here. 


\section{Results}
\label{sec:single_plant_case}

\Autoref{fig:learn_curve} displays the increasing trend of the earned reward while the A$2$C agent performed $300$K~interactions to learn the TVM task in the SP-Strawberry environments, which eventually took around $90$~minutes. 
\Autoref{tab:sp_mp_result} suggests that there is a strong correlation between the reward and the final OER, supporting the proposed design of reward function. 
Although the reward for~A2C dropped to $.052$ against the test set of unseen data, it was still significantly higher than the levels that other baseline methods could achieve.
Moreover, A$2$C agent led to over $18\%$ higher OER than other approaches. 
While the poorest performance of UDD indicates the low utility of the \emph{up} and \emph{down} actions, \autoref{fig:sp_st_acts_dist} reveals that A$2$C actually learned these ``seemingly weak'' actions to successfully combine with others.
\Autoref{fig:sp_st_acts} also visualizes the dynamic action choices with diversity. 

%
%
%

\autoref{tab:sp_mp_result} shows similar results from the MP~Globlet Vines scenarios. 
Though A$2$C still offers the best performance, the gap with other baselines has become slightly more marginal,
motivating creation of better models to generate reliable longer plans with more options of action in a larger field.  

\subsection{Real-World Transferability}
\label{fig:transfer_test}

We also investigate the realism of \mbox{DAVIS-Ag} since our ultimate goal is to assist in prototyping of active vision-enabled autonomous systems in actual plant environments.  
As a proxy, we follow previous protocols for synthetic datasets~\citep{SHVHLK21}, designing a ripe strawberry detection problem to test whether the performance is enhanced, if the detector is pre-trained with \mbox{DAVIS-Ag} and fine-tuned with real images. 

Specifically, a real strawberry-image dataset collected under natural field/light conditions in~\citep{LBBP22} was utilized along with the YOLOv$5$n object detector~\citep{G20}. 
We employed three different random splits---each consisting of $654$~and $159$~images in training and test sets, respectively---to investigate performance under different compositions of data. 
For pre-training, $2,565$~images from $9$~SP-Strawberry environments were employed as a training set, while $570$~images from two other environments were used as a validation set.   

\Autoref{tab:detection_results} indicates better performance of transfer learning (\mbox{DAVIS-Ag} $\rightarrow$ Real) compared to learning solely from the real training data (\mbox{Real-Only}) across most splits. 
This improvement hints the presence of visual elements shared by \mbox{DAVIS-Ag} and reality, demonstrating the feasibility of using the dataset for real-world prototyping.
Yet, further investigations into other crop types (such as~tomatoes and grapes), varying growth stages, and additional techniques (e.g.,~style transfer~\citep{MWGLBPS23}) may be needed to better warrant minimized sim-to-real gaps.  

\section{Conclusion \& Future Work}

We have introduced \mbox{DAVIS-Ag}, which, unlike previous agricultural datasets, features spatially dense image samples connected based on reachability to simulate motions of embodied agents.
It provides rich labels, such as bounding boxes and instance segmentation, to assist in learning for not only vision-related tasks but also active motion planning. 
Experiments evaluated baseline methods and demonstrated the realism of the dataset for prototyping real-world solutions. 
Our future work includes incorporating $3$D~data, more plant types, and real robot platforms in larger-scale simulations.

\section*{ACKNOWLEDGEMENT}
This work was partially supported by the grants from the NSF (OIA-2134901) and USDA-NIFA (USDA-020-67021-32855).

{\small
    \ifx\usenatbib\undefined%
	\bibliographystyle{IEEEtran}%
    \else%
    \bibliographystyle{IEEEtranN}%
    \fi
	\bibliography{CASE}

\begin{thebibliography}{46}
\providecommand{\natexlab}[1]{#1}
\providecommand{\url}[1]{#1}
\csname url@samestyle\endcsname
\providecommand{\newblock}{\relax}
\providecommand{\bibinfo}[2]{#2}
\providecommand{\BIBentrySTDinterwordspacing}{\spaceskip=0pt\relax}
\providecommand{\BIBentryALTinterwordstretchfactor}{4}
\providecommand{\BIBentryALTinterwordspacing}{\spaceskip=\fontdimen2\font plus
\BIBentryALTinterwordstretchfactor\fontdimen3\font minus \fontdimen4\font\relax}
\providecommand{\BIBforeignlanguage}[2]{{%
\expandafter\ifx\csname l@#1\endcsname\relax
\typeout{** WARNING: IEEEtranN.bst: No hyphenation pattern has been}%
\typeout{** loaded for the language `#1'. Using the pattern for}%
\typeout{** the default language instead.}%
\else
\language=\csname l@#1\endcsname
\fi
#2}}
\providecommand{\BIBdecl}{\relax}
\BIBdecl

\bibitem[Kootstra et~al.(2021)Kootstra, Wang, Blok, et~al.]{KWBHV21}
G.~Kootstra, X.~Wang, P.~M. Blok \emph{et~al.}, ``Selective harvesting robotics: current research, trends, and future directions,'' \emph{Current Robotics Reports}, 2021.

\bibitem[Choi et~al.(2022)Choi, Would, Salazar-Gomez, and Cielniak]{CWSC22}
T.~Choi, O.~Would, A.~Salazar-Gomez, and G.~Cielniak, ``Self-supervised representation learning for reliable robotic monitoring of fruit anomalies,'' in \emph{ICRA}, 2022.

\bibitem[Olenskyj et~al.(2022)Olenskyj, Sams, Fei, et~al.]{OSFSRBE22}
A.~G. Olenskyj, B.~S. Sams, Z.~Fei \emph{et~al.}, ``End-to-end deep learning for directly estimating grape yield from ground-based imagery,'' \emph{Comput. Electron. Agric.}, 2022.

\bibitem[Bajcsy(1988)]{B88}
R.~Bajcsy, ``Active perception,'' \emph{IEEE}, 1988.

\bibitem[Lehnert et~al.(2019)Lehnert, Tsai, Eriksson, and McCool]{LTEM19}
C.~Lehnert, D.~Tsai, A.~Eriksson, and C.~McCool, ``{3D} move to see: Multi-perspective visual servoing towards the next best view within unstructured and occluded environments,'' in \emph{IROS}, 2019.

\bibitem[van Essen et~al.(2022)van Essen, Harel, Kootstra, and Edan]{VHKE22}
R.~van Essen, B.~Harel, G.~Kootstra, and Y.~Edan, ``Dynamic viewpoint selection for sweet pepper maturity classification using online economic decisions,'' \emph{Applied Sciences}, 2022.

\bibitem[Zaenker et~al.(2021{\natexlab{a}})Zaenker, Smitt, McCool, and Bennewitz]{ZSMB21}
T.~Zaenker, C.~Smitt, C.~McCool, and M.~Bennewitz, ``Viewpoint planning for fruit size and position estimation,'' in \emph{IROS}, 2021.

\bibitem[Burusa et~al.(2022)Burusa, van Henten, and Kootstra]{BVK22}
A.~K. Burusa, E.~J. van Henten, and G.~Kootstra, ``Attention-driven active vision for efficient reconstruction of plants and targeted plant parts,'' \emph{arXiv}, 2022.

\bibitem[Zeng et~al.(2022{\natexlab{a}})Zeng, Zaenker, and Bennewitz]{ZZB22}
X.~Zeng, T.~Zaenker, and M.~Bennewitz, ``Deep reinforcement learning for next-best-view planning in agricultural applications,'' in \emph{ICRA}, 2022.

\bibitem[Menon et~al.(2022)Menon, Zaenker, and Bennewitz]{MZB22}
R.~Menon, T.~Zaenker, and M.~Bennewitz, ``Viewpoint planning based on shape completion for fruit mapping and reconstruction,'' \emph{arXiv}, 2022.

\bibitem[H{\"a}ni et~al.(2020)H{\"a}ni, Roy, and Isler]{HRI20}
N.~H{\"a}ni, P.~Roy, and V.~Isler, ``Minneapple: a benchmark dataset for apple detection and segmentation,'' \emph{IEEE Robotics and Automation Letters}, vol.~5, no.~2, pp. 852--858, 2020.

\bibitem[Yang et~al.(2019)Yang, Ren, Xu, Chen, Crandall, Parikh, and Batra]{YRXCCPB19}
J.~Yang, Z.~Ren, M.~Xu, X.~Chen, D.~J. Crandall, D.~Parikh, and D.~Batra, ``Embodied amodal recognition: Learning to move to perceive objects,'' in \emph{Proceedings of the IEEE/CVF International Conference on Computer Vision}, 2019.

\bibitem[Safronov et~al.(2021)Safronov, Piga, Colledanchise, and Natale]{SPCN21}
E.~Safronov, N.~Piga, M.~Colledanchise, and L.~Natale, ``Active perception for ambiguous objects classification,'' in \emph{IROS}, 2021.

\bibitem[Fang et~al.(2022)Fang, Liang, Wu, Xu, and Lim]{FLWXL22}
F.~Fang, W.~Liang, Y.~Wu, Q.~Xu, and J.-H. Lim, ``Self-supervised reinforcement learning for active object detection,'' \emph{RA-L}, 2022.

\bibitem[Cheng et~al.(2018)Cheng, Agarwal, and Fragkiadaki]{CAF18}
R.~Cheng, A.~Agarwal, and K.~Fragkiadaki, ``Reinforcement learning of active vision for manipulating objects under occlusions,'' in \emph{CoRL}, 2018.

\bibitem[Nilsson et~al.(2021)Nilsson, Pirinen, G{\"a}rtner, and Sminchisescu]{NPGS21}
D.~Nilsson, A.~Pirinen, E.~G{\"a}rtner, and C.~Sminchisescu, ``Embodied visual active learning for semantic segmentation,'' in \emph{AAAI}, 2021.

\bibitem[Zeng et~al.(2022{\natexlab{b}})Zeng, Li, Ran, Li, Gao, Li, He, Ye, et~al.]{ZLRLGLHY22}
J.~Zeng, Y.~Li, Y.~Ran, S.~Li, F.~Gao, L.~Li, S.~He, Q.~Ye \emph{et~al.}, ``Efficient view path planning for autonomous implicit reconstruction,'' \emph{arXiv}, 2022.

\bibitem[Ye et~al.(2018)Ye, Lin, Li, Zheng, and Yang]{YLLZY18}
X.~Ye, Z.~Lin, H.~Li, S.~Zheng, and Y.~Yang, ``Active object perceiver: Recognition-guided policy learning for object searching on mobile robots,'' in \emph{IROS}, 2018.

\bibitem[Schmid et~al.(2019)Schmid, Lauri, and Frintrop]{SLF19}
J.~F. Schmid, M.~Lauri, and S.~Frintrop, ``Explore, approach, and terminate: Evaluating subtasks in active visual object search based on deep reinforcement learning,'' in \emph{IROS}, 2019.

\bibitem[Hodan et~al.(2017)Hodan, Haluza, Obdr{\v{z}}{\'a}lek, Matas, Lourakis, and Zabulis]{HHOMLZ17}
T.~Hodan, P.~Haluza, {\v{S}}.~Obdr{\v{z}}{\'a}lek, J.~Matas, M.~Lourakis, and X.~Zabulis, ``{T-LESS}: An rgb-d dataset for 6d pose estimation of texture-less objects,'' in \emph{WACV}, 2017.

\bibitem[Singh et~al.(2014)Singh, Sha, Narayan, Achim, and Abbeel]{SSNAA14}
A.~Singh, J.~Sha, K.~S. Narayan, T.~Achim, and P.~Abbeel, ``Bigbird: A large-scale 3{D} database of object instances,'' in \emph{ICRA}, 2014.

\bibitem[Yang et~al.(2021)Yang, Gao, Li, and Waslander]{YGLW21}
J.~Yang, Y.~Gao, D.~Li, and S.~L. Waslander, ``{ROBI}: A multi-view dataset for reflective objects in robotic bin-picking,'' in \emph{IROS}, 2021.

\bibitem[Ammirato et~al.(2017)Ammirato, Poirson, Park, Ko{\v{s}}eck{\'a}, and Berg]{APPKB17}
P.~Ammirato, P.~Poirson, E.~Park, J.~Ko{\v{s}}eck{\'a}, and A.~C. Berg, ``A dataset for developing and benchmarking active vision,'' in \emph{ICRA}, 2017.

\bibitem[Zhao et~al.(2022)Zhao, Zhang, Wu, et~al.]{ZZWQL22}
Q.~Zhao, L.~Zhang, L.~Wu \emph{et~al.}, ``A real {3D} embodied dataset for robotic active visual learning,'' \emph{RA-L}, 2022.

\bibitem[Puig et~al.(2023)Puig, Undersander, Szot, et~al.]{AIHabitat3}
X.~Puig, E.~Undersander, A.~Szot \emph{et~al.}, ``Habitat 3.0: A co-habitat for humans, avatars and robots,'' \emph{arXiv}, 2023.

\bibitem[Ding et~al.(2023)Ding, Majcherczyk, Deshpande, Qi, Zhao, Madhivanan, and Sen]{DMDQZMS23}
W.~Ding, N.~Majcherczyk, M.~Deshpande, X.~Qi, D.~Zhao, R.~Madhivanan, and A.~Sen, ``Learning to view: Decision transformers for active object detection,'' in \emph{ICRA}, 2023.

\bibitem[Gibbs et~al.(2019)Gibbs, Pound, French, et~al.]{GPFWMP19}
J.~A. Gibbs, M.~P. Pound, A.~P. French \emph{et~al.}, ``Active vision and surface reconstruction for {3D} plant shoot modelling,'' \emph{TCBB}, 2019.

\bibitem[Rohmer et~al.(2013)Rohmer, Singh, and Freese]{RSF13}
E.~Rohmer, S.~P. Singh, and M.~Freese, ``{V-REP}: A versatile and scalable robot simulation framework,'' in \emph{IROS}, 2013.

\bibitem[Goslin and Mine(2004)]{GM04}
M.~Goslin and M.~R. Mine, ``The {Panda3D} graphics engine,'' \emph{Computer}, vol.~37, no.~10, pp. 112--114, 2004.

\bibitem[Quigley et~al.(2009)Quigley, Conley, Gerkey, Faust, Foote, Leibs, Wheeler, Ng, et~al.]{QCGFFLWN09}
M.~Quigley, K.~Conley, B.~Gerkey, J.~Faust, T.~Foote, J.~Leibs, R.~Wheeler, A.~Y. Ng \emph{et~al.}, ``{ROS}: an open-source robot operating system,'' in \emph{ICRA workshop}, 2009.

\bibitem[Koenig and Howard(2004)]{KH04}
N.~Koenig and A.~Howard, ``Design and use paradigms for gazebo, an open-source multi-robot simulator,'' in \emph{IROS}, 2004.

\bibitem[Chebrolu et~al.(2017)Chebrolu, Lottes, Schaefer, et~al.]{CLSWBS17}
N.~Chebrolu, P.~Lottes, A.~Schaefer \emph{et~al.}, ``Agricultural robot dataset for plant classification, localization and mapping on sugar beet fields,'' \emph{IJRR}, 2017.

\bibitem[Weyler et~al.(2023)Weyler, Magistri, Marks, et~al.]{WMMCSRCSB23}
J.~Weyler, F.~Magistri, E.~Marks \emph{et~al.}, ``Phenobench--a large dataset and benchmarks for semantic image interpretation in the agricultural domain,'' \emph{arXiv}, 2023.

\bibitem[Marzoa~Tanco et~al.(2023)Marzoa~Tanco, Trinidad~Barnech, Andrade, et~al.]{MTABLDT23}
M.~Marzoa~Tanco, G.~Trinidad~Barnech, F.~Andrade \emph{et~al.}, ``Magro dataset: A dataset for simultaneous localization and mapping in agricultural environments,'' \emph{IJRR}, 2023.

\bibitem[Polvara et~al.(2023)Polvara, Molina, Hroob, et~al.]{PMHPTGLTCH23}
R.~Polvara, S.~Molina, I.~Hroob \emph{et~al.}, ``Bacchus long-term ({BLT}) data set: Acquisition of the agricultural multimodal blt data set with automated robot deployment,'' \emph{JFR}, 2023.

\bibitem[Guevara et~al.(2023)Guevara, Joshi, Raja, et~al.]{GJRFBE23}
D.~Guevara, A.~Joshi, P.~Raja \emph{et~al.}, ``An open source simulation toolbox for annotation of images and point clouds in agricultural scenarios,'' in \emph{ISVC}, 2023.

\bibitem[Bailey(2019)]{B19}
B.~N. Bailey, ``Helios: A scalable {3D} plant and environmental biophysical modeling framework,'' \emph{Front. in Plant Sci.}, 2019.

\bibitem[Weber and Penn(1995)]{weber1995}
J.~Weber and J.~Penn, ``Creation and rendering of realistic trees,'' in \emph{SIGGRAPH}, 1995.

\bibitem[Kirk et~al.(2021)Kirk, Mangan, and Cielniak]{KMC21}
R.~Kirk, M.~Mangan, and G.~Cielniak, ``Robust counting of soft fruit through occlusions with re-identification,'' in \emph{ICVS}, 2021.

\bibitem[Zaenker et~al.(2021{\natexlab{b}})Zaenker, Lehnert, McCool, and Bennewitz]{ZLMB21}
T.~Zaenker, C.~Lehnert, C.~McCool, and M.~Bennewitz, ``Combining local and global viewpoint planning for fruit coverage,'' in \emph{ECMR}, 2021.

\bibitem[Polvara et~al.(2022)Polvara, Mellado, et~al.]{PMHCH22}
R.~Polvara, S.~M. Mellado \emph{et~al.}, ``Collection and evaluation of a long-term 4{D} agri-robotic dataset,'' \emph{arXiv}, 2022.

\bibitem[Mnih et~al.(2016)Mnih, Badia, Mirza, Graves, Lillicrap, Harley, Silver, and Kavukcuoglu]{MBMGLHSK16}
V.~Mnih, A.~P. Badia, M.~Mirza, A.~Graves, T.~Lillicrap, T.~Harley, D.~Silver, and K.~Kavukcuoglu, ``Asynchronous methods for deep reinforcement learning,'' in \emph{ICML}, 2016.

\bibitem[Lemsalu et~al.(2022)Lemsalu, Bloch, Backman, and Pastell]{LBBP22}
M.~Lemsalu, V.~Bloch, J.~Backman, and M.~Pastell, ``Real-time {CNN}-based computer vision system for open-field strawberry harvesting robot,'' \emph{IFAC-PapersOnLine}, 2022.

\bibitem[Shermeyer et~al.(2021)Shermeyer, Hossler, Van~Etten, et~al.]{SHVHLK21}
J.~Shermeyer, T.~Hossler, A.~Van~Etten \emph{et~al.}, ``Rareplanes: Synthetic data takes flight,'' in \emph{WACV}, 2021.

\bibitem[Jocher(2020)]{G20}
\BIBentryALTinterwordspacing
G.~Jocher, ``Yolov5 by ultralytics,'' 2020. [Online]. Available: \url{https://github.com/ultralytics/yolov5}
\BIBentrySTDinterwordspacing

\bibitem[Magistri et~al.(2023)Magistri, Weyler, et~al.]{MWGLBPS23}
F.~Magistri, J.~Weyler \emph{et~al.}, ``From one field to another---unsupervised domain adaptation for semantic segmentation in agricultural robotics,'' \emph{Comput. Electron. Agric.}, 2023.

\end{thebibliography}
}

\end{document}